# Latent Semantic Analysis Approach for Document Summarization Based on Word Embeddings

**Kamal Al-Sabahi[1], Zhang Zuping[1], Yang Kang[1]**

[1] School of Information Science and Engineering, Central South University

Hunan, Changsha 410083- China

{k.alsabahi, zpzhang, yk_ahead}@ csu.edu.cn

## *Abstract*

Since the amount of information on the internet is growing rapidly, it is not easy for a user to find relevant information for his/her query. To tackle this issue, much attention has been paid to Automatic Document Summarization. The key point in any successful document summarizer is a good document representation. The traditional approaches based on word overlapping mostly fail to produce that kind of representation. Word embedding, distributed representation of words, has shown an excellent performance that allows words to match on semantic level. Naively concatenating word embeddings makes the common word dominant which in turn diminish the representation quality. In this paper, we employ word embeddings to improve the weighting schemes for calculating the input matrix of Latent Semantic Analysis method. Two embedding-based weighting schemes are proposed and then combined to calculate the values of this matrix. The new weighting schemes are modified versions of the augment weight and the entropy frequency. The new schemes combine the strength of the traditional weighting schemes and word embedding. The proposed approach is experimentally evaluated on three well-known English datasets, DUC 2002, DUC 2004 and Multilingual 2015 Single-document Summarization for English. The proposed model performs comprehensively better compared to the state-of-the-art methods, by at least 1% ROUGE points, leading to a conclusion that it provides a better document representation and a better document summary as a result.

***Keywords:*** Word Embedding, Augment Weight, Entropy Frequency, Word2Vec, Document Summarization, Latent Semantic analysis

## 1. Introduction

The exponential growth of online text documents on the internet makes the need to compress and summarize text documents important and urgent. It is considered an effective solution for information overload. The text summarization technique is used to extract the most essential information from the original document and generate a simplified version of that document (Hu, Chen, and Chou 2017). A good summary should never contain repeated descriptions for the same piece of information. In other words, it is expected to have high diversity and minimum redundancy (Xiong and Ji 2016). Over the years, text summarization has seen advances in the sophistication of language processing techniques (Yao, Wan, and Xiao 2017). Several approaches have been proposed, ranging from simple position and word-frequency methods, to graph based and machine learning algorithms (Sankarasubramaniam, Ramanathan, and Ghosh 2014). The key point behind all these techniques is to find a representative subset of the original document such that the core essence of the document is contained in this subset from the semantic and conceptual standpoints (Sarkar 2016). Some of these techniques are based on Latent Semantic Analysis (LSA), where the document is represented as an input matrix *A*.

Latent Semantic Analysis (LSA) is a powerful unsupervised analytical tool. It is the most prominent algebraic learning algorithm used for Information Retrieval (Sarkar 2016). With a combination of statistical and algebraic methods, this method reveals the latent structure of words, among words, sentences or text through Singular Value Decomposition (SVD) algorithm. It produces measures of word-word, word-document and document-document relations that are well correlated with several human cognitive phenomena involving association or semantic similarity. The performance of LSA-based summarization algorithms depends on the quality of the document representation, the input matrix (Triantafillou et al. 2016, Al-Sabahi et al. 2018). Earlier unsupervised document representation approaches mostly rely on frequency and centrality. The assumption behind frequency-driven approaches is that the most important information will appear more frequently in the documents than less important detailed descriptions (Yao, Wan, and Xiao 2017). Traditional methods such as bag of words (BOW) and TF-IDF mostly fail to produce a good document representation. In addition, they need much processing and external resources. For example, in TF-IDF weighting scheme, terms only contribute to the score only if they match perfectly. In natural language, it is very common that different words are used to describe the same object in a text, so naive similarity measures between words cannot perfectly describe the content similarity (Han et al. 2016a). Recently, word embedding has successfully allowed words to match on the semantic level (Boom et al. 2015). Word embedding methods learn the continuous distributed vector representation of words with neural networks, which can capture the semantic and/or syntactic relationships. The great thing about word embedding is that it does not require a prior knowledge of natural language or external resources of structured semantic information; it just requires a large amount of unlabeled text data which used to create the semantic space (Kenter and Rijke 2015). The basic idea behind word embedding is that the embedding of each word represents its meaning. The challenge of using word embedding is in choosing a way of describing the distribution of word embeddings across the semantic space. The mean and the sum are rather poor ways of describing this distribution. It would be desirable to capture more features of the text, especially with respect to the terms that do not match.

### 1.1. Research Objectives

The main contribution of this work is proposing an unsupervised approach for extractive single-document summarization that combines the strength of word embedding with the strength of traditional weighting schemes such as Augment Weight (AW) and Entropy Frequency (EF). The ultimate goal is to improve the LSA-based algorithm to solve the automatic text summarization problem. Although word embedding has been applied to summarization task as a part of a language model, to our knowledge, we are the first to use the learned representation of word embedding to enhance the weighting scheme of the LSA input matrix. In summary, the objectives of this work are:

- Propose a novel local weighting scheme for the terms in a sentence, which is a modified version of the augment weight with word embeddings (EMBAW), section 3.2.1, b, 1.
- Propose a novel global weighting scheme for terms in the document, which is a modified version of the entropy frequency with word embeddings (EMBEF), section 3.2.1, b, 2.
- Compare the ROUGE results of our model and several baselines on the three datasets, section 5.4.

To verify the effectiveness of the proposed model, we conduct an extensive evaluation with several baselines on the two datasets. The evaluation results demonstrate that our model outperforms the state-of-the-art models on those datasets.

The rest of the paper is organized as follows. The related work is briefly described in section 2. In section 3, the proposed approach for summarizing documents is presented in details. The complexity analysis of the proposed model is introduced in section 4. Section 5 describes the experiments and results. Finally, we discuss the results and conclude in section 6.

## 2. Related work

Document summarization is one of the most difficult, though promising, application of Natural Language Processing (NLP). The researchers have been striving to utilize any advancement in NLP to create a more efficient summary. Extractive summarization techniques are the most common ones, whose basic idea is to extract important sentences from text documents, and then recombine them to form a summary (Wu et al. 2017). In this section, we briefly review the work higher relevant to the study of this paper, including: LSA based approaches, embedding-based approaches and deep learning-based ones.

### 2.1. LSA based models

LSA based document summarization approaches usually go through three main stages, Input Matrix Creation, SVD, and Sentence Selection (Al-Sabahi et al. 2018). Almost all the previous approaches perform the first two phases of LSA algorithm in the same way, where TF-IDF is used as a common weighting scheme. They differ in the way of selecting sentences for the summary, sentence selection algorithm. The most popular sentence selection algorithm which we will follow in this work is the one proposed by Steinberger and Jezek (Steinberger and Jezek 2004). In which they used both $V^T$ and $\Sigma$ matrixes for sentence selection process. The length of a sentence vector is determined by the concepts whose indexes are less than or equal to the number of dimensions in the new space. This dimension is given as a parameter to the algorithm. The singular values in $\Sigma$ matrix are used to give more importance to the concepts that are highly related to the text. The major weakness of the current LSA based summarization approaches is the using of the traditional weighting schemes which mostly fail to produce a good document representation.

Another LSA based study carried out by (Shen et al. 2014). In which, they tried to capture the important contextual information for latent semantic modeling by proposing a latent semantic model based on the convolutional neural network with convolution-pooling structure, called the convolutional latent semantic model (CLSM). The CLSM first presents each word within its context to a low-dimensional continuous feature vector, which directly captures the contextual features at the word n-gram level. Then, the CLSM discovers and aggregates only the salient semantic concepts to form a sentence level feature vector. After that, the sentence level feature vector is further fed to a regular feed-forward neural network, which performs a non-linear transformation, to extract high-level semantic information of the word sequence. However, this method was not basically proposed to solve document summarization problem.

### 2.2. Word Embedding-based models

Since traditional semantic sentence representations employ only WordNet or a corpus-based measure in a classic bag of words (Han et al. 2016b), there are many attempts to build sentence-level and document-level semantic information. In 2013, Mikolov, from Google, (Mikolov et al. 2013) proposed word2vec which was a major breakthrough in the direction of text representation. Nowadays, word embedding is taking apart in many NLP applications. Furthermore, word embedding substitutes the external semantic knowledge and make human

"feature engineering" unnecessary (Kenter and Rijke 2015). In the context of document summarization, some studies used the sum of trained word embeddings to represent sentences or documents. Recently, the researchers are striving to go from word embeddings into sentence-level and document-level embeddings, because they are more suitable for the task.

A sentence embedding based on a large-scale training set of paraphrase pairs was built by (Wieting et al. 2015). They used a neural network model to optimize the word embeddings such that the cosine similarity of the sentence embeddings for the pair is maximized. In addition, they used a paraphrase pair and word embeddings to build the sentence embeddings. Document-level similarity was used by (Kobayashi, Noguchi, and Yatsuka 2015) to summarize the documents. They proposed a submodular function based on word embeddings. The objective function of their model is calculated based on the k-nearest neighbor distance on embedding distributions. Then, a modified greedy algorithm is used to select sentences for the summary. A centroid-based method was proposed by (Rossiello, Basile, and Semeraro 2017) that utilizes word2vec to find the centroid by summing the embeddings of the top ranked words which has TF-IDF greater than the topic threshold. The score of the sentence is the summation of the vectors of its words.

It is worth mentioning that using the word embeddings alone may make the high frequent unimportant words dominant which in turn diminish the representation accuracy. In this work, we proposed a new local and global weighting schemes that combine the traditional weighting schemes with word embeddings to improve the performance of LSA on the document summarization task.

### 2.3. Deep learning-based approaches

Deep neural networks have been used for both abstractive summarization and extractive summarization (Al-Sabahi, Zuping, and Nadher 2018). For extractive summarization, (Yousefi-Azar and Hamey 2017) proposed extractive query-oriented single-document summarization using a deep auto- encoder to compute a feature space from the term-frequency (tf) input. They developed a local word representation in which each vocabulary is designed to build the input representations of the document sentences. Then, a random noise is added to the word representation vector, affecting both the inputs and outputs of the auto-encoder.

In 2015, (Rush, Chopra, and Weston 2015) published an encoder-decoder model in which the encoder is a convolutional network and the decoder is a feedforward neural network language model. They enhanced the convolutional encoder by integrating it with attention model. As the convolutional encoder need a fix number of features, they used a bag of n-grams model. That means they ignore the overall sequence order while generating the hidden representation. They only used the first sentence of each news article to generate its title.

A Recurrent Neural Network (RNN) based sequence model for extractive summarization of documents presented by (Nallapati, Zhai, and Zhou 2016). In which, they treated extractive summarization as a sequence classification problem. They used neural networks for the sentential extractive summarization of single documents. In this model, each sentence is visited sequentially as it appears in the original document and a binary decision is made in terms of whether or not it should be included in the summary. Another study carried out by (Cao et al. 2016) tried to learn the distributed representations for sentences by applying an attention mechanism, which used to learn query relevance ranking and sentence saliency ranking simultaneously. A recent work proposed by (See, Liu, and Manning 2017) in which they augmented the standard sequence-to-sequence attentional model in two orthogonal ways. In first way, they used a hybrid pointer-generator network that can copy words from the source text via pointing. In the second one, they used the coverage to keep track of what has been summarized so far.

After exploring some deep learning-based model, it is worth mentioning two things: first, most of the previous deep learning-based models are supervised approaches, which need a huge amount of labeled data. Creating a suitable labeled data for text summarization is very challenging since the past studies have shown that, human summarizers tend to agree, approximately, only about 60% of the times, and in only, 82% of the cases humans agreed with their own judgment (Fuentes et al. 2005, Al-Sabahi et al. 2018). In our work, we utilize word embeddings along with some traditional methods to build a robust unsupervised model. Second, most of the recent work has focused on headline generation tasks that means reducing one or two sentences to a single headline (See, Liu, and Manning 2017).

## 3. The proposed model

To get the summary, defined in Definition 1, a new LSA-based algorithm is proposed. Instead of using the traditional methods to calculate the values of the input matrix (*A*), a word embedding based method is used. The new method is based on modified versions of the augment weight as a local weight and the entropy frequency as a global weight. The enhanced weighting scheme helps LSA algorithm to provide summaries with reasonable quality compared to existing models. As shown in **Fig.** 1, the proposed model consists of two main stages: finding word vector and LSA algorithm implementation. The subsections 3.1 and 3.2 explain these stages in more details.

**Definition 1**. Given a document *D* consisting of a sequence of sentences $(s_1, s_2 ..., s_n)$ and a set of words $(w_1, w_2, ..., w_m)$. Sentence extraction aims to create a summary from *D* by selecting a subset of *M* sentences (where *M* < *n*) (Cheng and Lapata 2016).

To set the scene of this work, we begin with a brief overview about the traditional weighting schemes limitations in finding a good document representation. For example, according to traditional vector representation and after removing stop-words, the following two similar, but word-different, sentences in one document: "Obama speaks to the media in Illinois" and "The President greets the press in Chicago" will have a maximum simplex distance, although their true distance is small (Kusner et al. 2015). Moreover, term frequency (TF) is an essential part in almost all the traditional methods. Its value is basically calculated as the number of occurrences of a specific term in a sentence or a document. Depending on the exact matching, TF mostly fail to give a perfect representation since the writer usually use different words to describe the same thing. **Table 1** shows the term frequency for all the terms in the two sentences in the previous example, excluding stop words. Although the two sentences are very similar, TF could not capture the semantic similarity between them. It is worth mentioning that we could use an external lexical database to overcome this to some extent, but this has some serious issues such as, hard accurate word similarity, missing new words, and needs a lot of human feature engineering.

**Table 1**. Term frequency for $s_1$ and $s_2$ in the above example calculated as the number of occurrences of each term in the sentences

|       | Obama | speaks | media | Illinois | president | greets | press | Chicago |
|-------|-------|--------|-------|----------|-----------|--------|-------|---------|
| $s_1$ | 1     | 1      | 1     | 1        | 0         | 0      | 0     | 0       |
| $s_2$ | 0     | 0      | 0     | 0        | 1         | 1      | 1     | 1       |

To solve this issue, we replace the term frequency, TF, with a new formula that able to work on the semantic level, Equation 7. The basic idea behind the new formula is replacing the term frequency with the cosine similarity between the word vector of the term and the ones of every word in the sentence. **Table 2** shows the similarity between terms and sentences in the previous example calculated using the proposed formula, Equation 7. From the results in **Table 2**, it is notable that the new formula captures more information and gives a better representation, so it can be used as a replacement of the traditional term frequency in many weighting schemes. In this work, we use the new formula to improve augment weight and the entropy frequency as discussed in section 3.2.

**Table 2.** The term-sentence similarity matrix for sentences in the above example calculated using Equation 7

|       | Obama | speaks | media | Illinois | president | greets | press | Chicago |
|-------|-------|--------|-------|----------|-----------|--------|-------|---------|
| $s_1$ | 1.603 | 1.269  | 1.283 | 1.548    | 0.328     | 0.749  | 1.013 | 1.279   |
| $s_2$ | 0.773 | 1.028  | 0.860 | 0.708    | 1.239     | 1.346  | 1.363 | 1.109   |

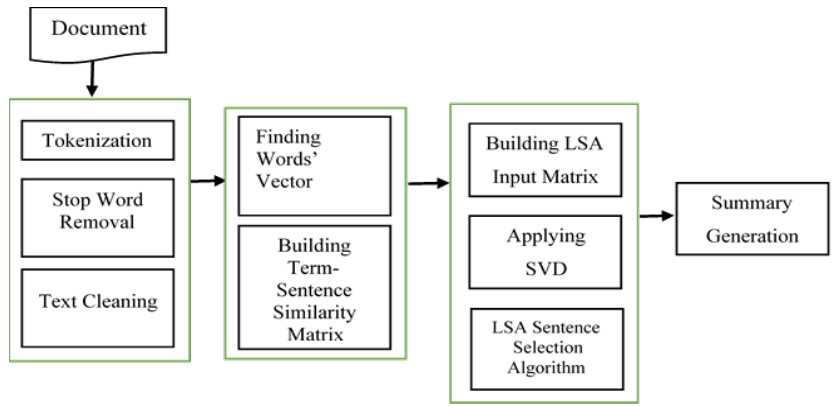

**Fig. 1.** The Proposed Model Architecture

### 3.1. Finding word vector

Learning word embedding is entirely unsupervised. Word2vec learns a vector representation for each word using a simple neural network language model. The network architecture, the skip-gram model, consists of an input layer, a projection layer, and an output layer. It uses this simple architecture to predict nearby words. Each word vector is trained to maximize the log probability of neighboring words in a very large corpus (Kusner et al. 2015). The ability to train on very large unlabeled datasets allows the model to learn complex word relationships.

**Definition 2.** For any two embedding distributions *A* and *B*, when *A* is similar to *B*, each embedding in *A* should be near to some embedding in *B*.

According to Definition 2, we can conclude that discovering such a complex relationship between the distributed representation of words compensates some of the key weaknesses of bag-of-words models (Kim and Lee 2016). In this work, we will use two freely available pre-trained word embeddings, Google word2vec and GloVe.

### 3.1.1. Google word2vec Model[1]

Pre-trained Google News corpus (GoogleNewsvectors-negative300.bin.gz) with (about 100 billion words) which has an embedding for 3 million words/phrases 300-dimension English word vectors, trained using the approach in (Mikolov et al. 2013). Words that are not present in the pre-trained word2vec model are dropped but kept for nBOW baseline.

### 3.1.2. GloVe Model[2]:

GloVe (Pennington, Socher, and Manning 2014) is an unsupervised learning algorithm for obtaining vector representation for words. Training is performed on aggregated global word-word co-occurrence statistics from a corpus. In this work, we use the one trained on Common Crawl (840B tokens, 2.2M vocab, cased, 300d vectors, 2.03 GB download): glove.840B.300d.zip.

## 3.2. LSA algorithm

The summarization algorithms that are based on LSA usually include three main steps: creating the input matrix (sentence - term matrix), applying SVD to matrix and selecting the sentences for the summary.

### 3.2.1. Building input matrix

The input document is represented as an $m \times n$ matrix ($A$). Each row in the matrix $A$ represents a term and each column represents a sentence. The cell value $a(w_i, s_j)$ represents the importance of term $w_i$ in sentence $s_j$. When forming the input matrix, $A$, a variety of well-known term weighting schemes can be used. There are two common types: a local weight based on the frequency within the sentence and a global weight based on a term's frequency throughout the document (Kontostathis 2007). In this work, as in Definition 3, the entry $a(w_i, s_j)$ is obtained by multiplying the local weight $L(w_i, s_j)$ for each term by the global weight $G(w_i)$.

**Definition 3**: Let the input document ($D$) represented as an $m \times n$ matrix, where $A[j] = [a(w_1, s_j), a(w_2, s_j), …, a(w_i, s_j)]$, $a(w_i, s_j)$ is calculated using Equation 1:

$$a(w_i, s_j) = L(w_i, s_j) \times G(w_i) \qquad (1)$$

, where: $L(w_i, s_j)$ is the Local Weight for term $w_i$ in sentence $s_j$ and $G(w_i)$ is the Global Weight for term $w_i$ in the whole document.

There are different weighting schemes to calculate the local weight and the global weight for each term. The traditional ways for calculating these weights are very difficult because they must pass through a pre-processing step and they depend heavily on the exact matching. In this work, we propose a modified version of the augment weight as a local weight, Definition 4, and a modified version of the entropy frequency as a global weight, Definition 5. The following subsections describe two types of weighting schemes combinations, a traditional one and a word embedding-based one, as follows:

### a) Traditional Weighting Scheme (AWEF)

---

[1] https://code.google.com/archive/p/word2vec/
[2] http://nlp.stanford.edu/projects/glove/

The combination of the augment weight and the entropy frequency is a traditional way for building the input matrix. In this work, we used the following equations (Manning, Raghavan, and Schütze 2008) to implement this weighting schemes used as a baseline:

**Augment Weight (AW):** The augment weight is calculated by Equation (2):

$$L(t_{ij}) = 0.5 + 0.5 \times (\frac{tf_{ij}}{tf_{\max}}) \qquad (2)$$

Where $tf_{ij}$ refers to the number of times that the $i^{th}$ term occurs in the $j^{th}$ sentence and $tf_{max}$ refers to the frequency of the most frequently occurring term in the $j^{th}$ sentence.

**Entropy Frequency (*EF*)**: The Entropy Frequency is calculated by Equations (3) & (4):

$$G(t_{ij}) = 1 + \sum \frac{P_{ij} \log_2 P_{ij}}{\log_2 n} \qquad (3)$$

, where

$$P_{ij} = \frac{tf_{ij}}{gf_i} \qquad (4)$$

Where $tf_{ij}$ is the number of times that the the $i^{th}$ term occurs in the $j^{th}$ sentence, $n$ is the total number of sentences, and $gf_i$ is the number of times that the $i^{th}$ term occurs in the entire document.

**b) Embedding-Based Weighting Scheme (EMBAWEF)**

Two embedding-based weighting schemes are proposed. The basic idea behind the new schemes is replacing the term frequency with the cosine similarity between the word vector of the term and the ones of every word in the sentence.

**Embedding-Based Augment Weight (EMBAW):** Definition 4 represents the embedding-based augment weight and **Algorithm 1** shows how it is calculated.

**Definition 4.** For the input document (*D*) with *n* sentences and *m* terms, let $D = (s_1, s_2, \ldots, s_n)$ where $s_j$ (1≤*j*≤*n*) denotes the $j^{th}$ sentence, *W* is a set of all terms in the document, and $V = (v_{w_1}, v_{w_2}, \ldots, v_{w_m})$ where $v_{w_i}$ ($1 \leq i \leq m$) refers to the word vector of the term $w_i$. Let $L(w_i, s_j)$ be the local weight for term $w_i$ in sentence $s_j$. For each $w_i$ ($1 \leq i \leq m$), the embedding-based augment weight for $s_j$ is calculated by Equation (5):

$$L(w_i, s_j) = 0.5 + 0.5 \times (\frac{TermSentSim(w_i, s_j)}{TermSentSim_{\max j}}) \qquad (5)$$

, where $TermSentSim(w_i, s_j)$ refers to the similarity score of the term $w_i$ with sentence $s_j$, calculated using Equation 6, and $TermSentSim_{maxj}$ refers to the similarity score of the term that has the maximum similarity with sentence $s_j$ calculated using Equation 7. **Fig. 2** shows the way of calculating $TermSentSim$ for the term $w_i$ with respect to sentence $s_j$.

$$TermSentSim(w_i, s_j) = \sum_{w' \in s_j} CosineSim(v_{w_i}, v_{w'}) \qquad (6)$$

, where $CosineSim(v_{w_i}, v_{w'})$ denotes the cosine similarity between word vector of term $w_i$ with respect to the ones of every term in sentence $s_j$.

$$TermSentSim_{maxj} = max_{w \in W} TermSentSim(w, s_j) \qquad (7)$$

**Embedding-Based Entropy Frequency (EMBEF):** The embedding-based Entropy frequency is defined in Definition 5. **Algorithm 2** shows how it is calculated.

**Definition 5.** Let $G(w_i)$ is the global weight for the term ($w_i$) in the entire document $D$. For each $w_i (1 \leq i \leq m)$, the Embedding-Based Entropy Frequency for the term $w_i$ is calculated by Equations (8) & (9):

$$G(w_i) = 1 + \sum_{j \in n} \frac{P(w_i, s_j) \log_2 P(w_i, s_j)}{\log_2 n} \qquad (8)$$

, where $n$ is the number of sentences in the document,

$$P(w_i, s_j) = \frac{TermSentSim(w_i, s_j)}{TermDocSim(w_i, D)} \qquad (9)$$

, where $TermSentSim(w_i, s_j)$ refers to the similarity between term $w_i$ and sentence $s_j$, calculated using Equation 7, and $TermDocSim(w_i, D)$ refers to the similarity score of term $w_i$ with respect to the entire document calculated using Equation 10.

$$TermDocSim(w_i, D) = \sum_{w'' \in D} CosineSim(v_{w_i}, v_{w''}) \qquad (10)$$

, where $CosineSim(v_{w_i}, v_{w''})$ denotes the cosine similarity between the word vector of term $w_i$ with respect to the ones of every term in the entire document $D$.

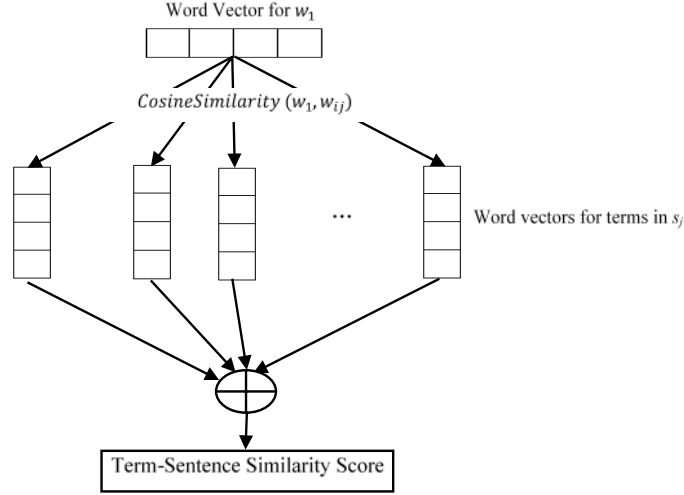

**Fig. 2.** Calculating Term-Sentence Similarity Scores

### 3.2.2. Singular Value Decomposition (SVD)

After building the input matrix $A$, SVD is applied, which is an algebraic method that can identify the relationships between terms and sentences (Lee et al. 2009).

$$A = U \Sigma V^T \qquad (11)$$

Equation 11 shows that $U$, $\Sigma$ and $V^T$ are the three matrices of SVD. $U$ is an $m \times n$ matrix which represents term by concept, $\Sigma$ represents the scaling values. The magnitude of singular values in $\Sigma$ suggests the degree of importance of the concepts. $V^T$ represents concept-by-sentence. In $V^T$ matrix, the columns represent the sentences and the rows represent the concepts.

The row order indicates the importance of the concepts, so the first row represents the most important concept. A higher cell value indicates that the sentence is more related to the concept. Once the term by sentence matrix ($A$) is constructed, we used *scipy.sparse.linalg* python library to apply SVD on the input matrix $A$.

### 3.2.3. The sentence selection algorithm

In this work, we implemented the popular algorithm proposed by (Steinberger and Jezek 2004) which proposes some improvements over some previous excellent work, as we mentioned in section 2.1.

## 4. Complexity Analysis

In this section, the proposed approach algorithms are represented along with their complexity analysis. **Algorithm 1** is used to calculate the embedding-based augment weight (EMAW). The algorithm finds the word vector for a word and compares it with the vectors of every word in the document. Let $|W|$ be the number of words in the document, $|S|$ is the number of sentences in the document and $|s_i|$ is the number of words in the longest sentence. The time complexity of this algorithm is determined by the most inner for loop, lines 5-8, that has the time complexity of $O(|W||S||s_i|)$; where $|S||s_i|$ is roughly equal $|W|$, the overall time complexity of **Algorithm 1** is $O(|W|^2)$. Comparing this algorithm with the traditional ones (AW), mentioned in section 3.2.1.a, will lead to a conclusion that they have the same complexity. The only difference is that the proposed algorithm needs to find the word vector from a lookup-table which takes $O(1)$ for each word, so the overall time complexity remains the same.

**Embedding Augment Weight Algorithm**
**Input**: a set of all terms in the document $W$, a set of sentences $S = (s_1, s_2, ..., s_j)$, a set of word vectors $V_w = (v_{w_1}, v_{w_2}, ..., v_{w_m})$.
**Output**: Embedding Augment Weight (*EMBAW*)

1   **For** each word $w$ in $W$ **do**
2     $v_w = w2v(w)$
3     **For** each sentence $s_i$ in $S$ **do**
4       $TermSim := 0$
5       **For** each word $w'$ in sentence $s_j$ **do**
6         $v_{w'} = w2v(w')$
7         $TermSim := TermSim + (CosineSimilarity(v_w, v_{w'}))$
8       **End For**
9       $TermSentSim[w, s_j] := TermSim$
10     **End For**
11   **End For**
12   $TermSentSim_{maxj} := max(TermSentSim, axis = 0)$
13   **For** each word $w$ in $W$ **do**
14     **For** each sentence $s_j$ in $S$ **do**
15       $EMBAW[w, s_j] := (0.5 + 0.5 \times TermSentSim[w, s_j]/TermSentSim_{maxj}[s_j])$
16     **End For**
17   **End For**

**Algorithm 1.** Pseudocode for calculating Embedding-Based Augment weight (EMBAW).

**Algorithm 2** is used to calculate the Embedding Entropy weight EMDEF. The term sentence similarity matrix (*TermSentSim*), calculated by **Algorithm 1**, is fed as input to **Algorithm 2**. The time complexity of this algorithm depends on the execution time of the most inner For loop, lines 4-8, that have the time complexity bounded to $O(|W||S|)$. This has the same complexity as the algorithm calculating the tradition Entropy Frequency (EF), mentioned in section 3.2.1.a.

|  | **Embedding Entropy Frequency Algorithm** |
|---|---|
|  | **Input** : a set of all words in the documents *W*, a set of sentences $S = (s_1, s_2, s_3, ..., s_j)$, and Term-Sentence Similarity Matrix $TermSentSim$ |
|  | **Output**: Embedding Entropy Frequency (*EMBEF*) |
| 1 | $TermDocSim[w, D] := Sum(TermSentSim, axis = 1)$ |
| 2 | **For** each word *w* in *W* **do** |
| 3 |    **For** each sentence $s_i$ in *S* **do** |
| 4 |       $P[w] := TermSentSim[w, s_j] / TermDocSim[w, D]$ |
| 5 |       **if** $P[w] > 0$ |
| 6 |          $Plg[w].append(P[w] \times log_2 P[w])$ |
| 7 |       **else** |
| 8 |          $Plg[w].append(0)$ |
| 9 |       **End if** |
| 10 |    **End For** |
| 11 | **End For** |
| 12 | **For** each word *w* in *W* **do** |
| 13 |    $EMBEF[w] := 1 + Sum(Plg[w] / log_2 |S|)$ |
| 14 | **End For** |

**Algorithm 2.** Pseudocode for calculating Embedding-Based Entropy Frequency (EMDEF).

## 5. The experiments and analysis

We compare the proposed approach, EMBAWEF, with the baselines models on two well-known datasets.

### 5.1. Datasets:

The proposed LSA based algorithm is evaluated on the three well-known datasets, DUC 2002, DUC 2004 and Multilingual 2015 Single-document Summarization (MSS 2015) (Giannakopoulos et al. 2015). The first dataset contains 567 news articles belonging to 59 different clusters of various news topics, and the corresponding 100-word manual summaries generated for each of these documents (single-document summarization), or the 100-word summaries generated for each of the 59 document clusters formed on the same dataset (multi-document summarization). In this work, the single-document summarization tasks are used. The other dataset is DUC 2004. This dataset consists of five tasks. Task 1 consists of 500 newspaper and newswire articles. Each document includes four very short gold standard summaries with a maximum length of 75 characters used for evaluation. Task 2 was of 50 clusters of related documents, each of which contains 10 documents. Each cluster of documents also includes four gold summaries with a maximum length of 665 characters (about 100 words) for each summary including spaces and punctuation. To check the effectiveness of the proposed method even further, Multilingual 2015 Single-document Summarization (MSS)

(Giannakopoulos et al. 2015) task is used. In MSS 2015 [3] task, single-document summaries are generated for some selected Wikipedia articles with at least one out of 38 languages defined by the organizers of the task. For each one of the 38 languages, there are 30 documents. In this work, the proposed method is evaluated for English. ROUGE-1 and ROUGE-2[4] are adopted for evaluation in this experiment.

### 5.2. Baselines:

To evaluate the proposed approaches, extensive comparison with multiple abstractive and extractive baselines are done. However, there are so many approaches to the text summarization problem, for comparison, we choose the ones that are comparable to our work on the three datasets as follows:

#### 5.2.1. On DUC-2002:

- Leading sentences (Lead-3): which simply produces the leading three sentences of the document as a summary. This model serves as a baseline on DUC 2002 dataset.
- Recurrent Neural Network based model (SummaRuNNer) proposed by (Nallapati, Zhai, and Zhou 2016) is used as a baseline on DUC 2002, mentioned in section 2.3.
- In addition, the well-known state-of-the-art (TF-IDF) shown a competitive performance on the DUC 2002 single document summarization task is used as a baseline.
- In addition, the extractive model proposed by (Cheng and Lapata 2016) was used as a baseline on this dataset.

#### 5.2.2. On DUC-2004:

- From the DUC-2004 single-document task, we include the PREFIX baseline that simply returns the first 75 characters of the input as a headline.
- Neural attention-based model (ABS+) proposed by (Rush, Chopra, and Weston 2015) is used as a baseline on DUC 2004, mentioned in section 2.3.
- We also report the TOPIARY system, which achieved the best performance in DUC 2004 shared task.
- For the DUC-2004 multi-document task, we used LEAD that simply chooses the first 100 words from the most recent article in each cluster. Moreover, to ensure a fair comparison, we compare our model with three popular models applied on this dataset. The first is an RNN-based model proposed by (Cao et al. 2015) that used RNN for learning sentence embeddings. The second is a centroid-based method (C SKIP) for text summarization that exploits the compositional capabilities of word embeddings proposed by (Rossiello, Basile, and Semeraro 2017). The third one is a neural graph-based model by (Yasunaga et al. 2017).

#### 5.2.3. On Multiling MSS 2015:

- The baselines used for comparison on Multiling MSS 2015 are the BEST and The WORST scores obtained by the 23 participating systems. In addition, the centroid-based model (C W2V) (Rossiello, Basile, and Semeraro 2017) is used as a baseline on this dataset for English.

---

[3] http://multiling.iit.demokritos.gr/pages/view/1532/taskmss-single-document-summarization-data-and-information

[4] ROUGE-1.5.5 with options -n 2 -2 4 -u -x -m

AWEF also serves as a baseline for the word embedding-based proposed model to measure the impact of using word embeddings while building LSA input matrix. It is worth mentioning that during implementation of the traditional TF-IDF and AWEF models, we employ the same LSA based selection algorithm with the same settings for selecting sentences for the summary.

### 5.3. Evaluation

In this work, the ROUGE (Recall-Oriented Understudy for Gisty Evaluation) metrics (Lin 2004) are used for the automatic evaluation of the generated summaries. The ROUGE metrics are based on the comparison of n-grams between the summary to be evaluated and one or several human written reference summaries. All the experiments described in this work were evaluated using the ROUGE Toolkit and the pyrouge package. We choose ROUGE-1, ROUGE-2, and ROUGE-L to evaluate this work.

**Remark 2**: To ensure that the recall-only evaluation will be unbiased to length, we use the ''-l 100'' and "-l b75" options in ROUGE to truncate longer summaries in DUC 2002 and DUC 2004 respectively.

### 5.4. Experimental results

All the experiments described in this work were evaluated using the ROUGE Toolkit[5] and the pyrouge package[6]. ROUGE-1, ROUGE-2, and ROUGE-L were applied. We compared our proposed model EMBAWEF with several baselines, mentioned in section 5.2. Furthermore, we implemented two LSA-based models as baselines for comparison. The first method is a combination of the traditional Augment weight and Entropy frequency (AWEF) as local and global weight respectively implemented and evaluated on the three datasets, DUC 2002, DUC 2004 and MSS 2015. The second one is the TF-IDF baseline implemented on the two datasets original documents. The SVD was applied on the matrix ($A$). After getting the three SVD matrices, the sentence selection algorithm of (Steinberger and Jezek 2004) was applied to select sentences for the summary. The output of the evaluation was compared to the human-generated summaries in these datasets using the ROUGE Toolkit. The evaluation results, shown in **Table 3**, **Table 4**, **Table 5, Table 6** and **Fig. 3**, assert that the proposed model achieves promising results. From the obtained results, we can make the following observations:

- **Table 3** and **Fig. 3** show that the obtained results for ROUGE-1, ROUGE-2, and ROUGE-L using EMBAWEF, AWEF, TF-IDF, and LEAD-3 indicate that our proposed method EMBAWEF performs the best for all ROUGE metrics used in this experiment on DUC 2002 dataset. This asserts that using word embedding in the weighting scheme leads to an increase in the effectiveness of detecting semantically similar sentences, compared to traditional methods. Furthermore, an important implication of the obtained results is that word embedding has come to a level where it can be employed in a generic approach for producing features that can be used to yield state-of-the-art performance on the text summarization task.

---

[5] http://www.berouge.com/Pages/default.aspx  ROUGE-1.5.5 with options: -n 2 -m -u -c 95 -r 1000 -f A -p 0.5 -t 0

[6] https://pypi.python.org/pypi/pyrouge/0.1.3

- The other proposed method AWEF performed better than the baselines LEAD-3 and TF-IDF. These results imply that combining AW and EF as a weighting scheme can capture more complex meanings than the other traditional combination of weighting schemes but it could not beat SummaRuNNer (Nallapati, Zhai, and Zhou 2016). On the other hand, using TF-IDF as a weighting scheme to calculate the input matrix performed the worst for all ROUGE metrics used in this work, **Table 3**. A possible reason is that a wide variety of expressions by users made it difficult to calculate similarities. In the case of ROUGE-2, LEAD-3 performed better than AWEF. According to (Lin and Hovy 2003), the higher order ROUGE-N is worse than ROUGE-1 since it tends to score grammaticality rather than content.

**Table 3.** The performance comparison of the proposed models using Word2Vec with respect to the baselines models on DUC 2002

| Model | Word2Vec on DUC 2002 | | |
|---|---|---|---|
| | ROUGE-1 | ROUGE-2 | ROUGE-L |
| EMBAWEF | **51.17%** | **23.43%** | **46.86%** |
| AWEF | 45.13% | 18.15% | 41.06% |
| TF-IDF | 43.57% | 17.46% | 39.15% |
| LEAD-3 | 43.60% | 21.00% | 40.20% |
| SummaRuNNer (Nallapati, Zhai, and Zhou 2017) | 47.36% | 22.10% | 42.03% |
| Cheng et al '16 (Cheng and Lapata 2016) | 47.40% | 23.00% | 43.50% |

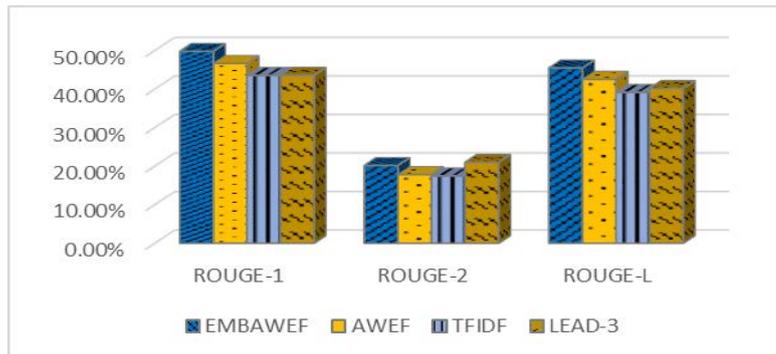

**Fig. 3.** The performance comparison of the proposed models using GloVe with respect to the baselines models on DUC 2002

- In the case of DUC 2004, **Table 4**, the results assert that the proposed models, EMBAWEF, outperforms all the non-deep learning based baselines in the term of all ROUGE metrics used in this experiment, but it could not beat the deep learning model (Rush, Chopra, and Weston 2015). There are two possible reasons for this. The first one is that the ABS+ model has been trained on a very huge dataset, annotated Gigaword dataset (Napoles, Gormley, and Durme 2012). The model is headline-generation which turn to achieve a higher ROUGE score, but they usually fail when they are asked to generate a longer summary (Paulus, Xiong, and Socher 2017) .However, the proposed model achieves a competitive performance. The second one is that the attentional encoder-decoder models, such as (Rush, Chopra, and Weston 2015), were able to generate short abstractive summaries with high ROUGE scores. While ROUGE measures the n-gram overlap between the generated summary and a reference one, summaries with high ROUGE scores are not necessarily the more readable ones. One potential issue of generative summarization models is that optimizing for a specific discrete metric like ROUGE does not guarantee an increase in quality and readability of the generated summary (Paulus, Xiong, and Socher 2017,

Liu et al.). This justifies the fact that our model could not beat the abstractive baseline, ABS+ (Rush, Chopra, and Weston 2015), for short summaries, as in **Table 4**.

**Table 4.** The performance comparison of the proposed models using Word2Vec with respect to the baselines models on DUC 2004 Single-Document Summarization

| Model | Word2Vec on DUC 2004 | | |
|---|---|---|---|
| | ROUGE-1 | ROUGE-2 | ROUGE-L |
| Ours (EMBAWEF) | 27.40% | 7.41% | 22.95% |
| AWEF | 25.68% | 6.39% | 21.32% |
| TF-IDF | 23.51% | 4.03% | 19.69% |
| PREFIX | 21.43% | 6.04% | 17.45% |
| TOPIARY | 25.12% | 6.46% | 20.12% |
| ABS+ (Rush, Chopra, and Weston 2015) | 28.18% | 8.49% | 23.81% |
| RAS-Elman (Chopra, Auli, and Rush 2016) | **28.97%** | 8.26% | 24.06% |
| words-lvt2k-1sent (Nallapati et al. 2016) | 28.35% | **9.46%** | **24.59%** |

- To provide a fair comparison, we applied the model on DUC-2004 multi-document summarization tasks 2. This task has summaries with an average length of 100 words (665 characters) instead of 75 characters for DUC-2004 single-document summarization task. As shown in **Table 5**, the proposed model performs well and outperforms all the baselines, which proves the feasibility and the superiority of the proposed model.

**Table 5.** The performance of the proposed model with respect to the baselines on DUC-2004 Multi-document Summarization Task 2.

| Model | Word2Vec on DUC 2004 | |
|---|---|---|
| | ROUGE -1 | ROUGE -2 |
| Ours (EMBAWEF) | **40.23%** | **10.41%** |
| LEAD | 32.42% | 6.42% |
| C SKIP (Rossiello, Basile, and Semeraro 2017) | 38.81% | 9.97% |
| RNN (Cao et al. 2015) | 38.78% | 9.86% |
| GRU+GCN (Yasunaga et al. 2017) | 38.23% | 9.48% |

- From the results in **Table 6**, it can be observed that our proposed approach (EMBAWEF) achieves better performance on MSS 2015 with respect to the worst and the best scores on this dataset. In addition, it outperformed the centroid-based model proposed by (Rossiello, Basile, and Semeraro 2017) used as a baseline in this work.
- **Fig. 4** demonstrates that the model with word2vec has outperformed the one with GloVe.

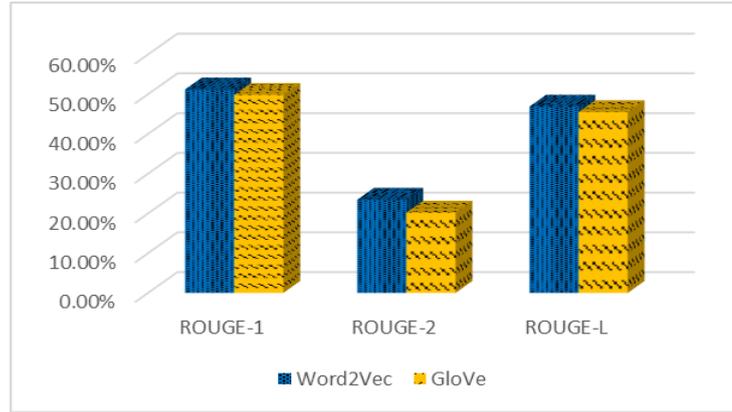

**Fig. 4.** The performance comparison of the EMBAWEF model using GloVe and Word2Vec on DUC 2002

- Finally, our proposed model, EMBAWEF, obtained good results competing with the state-of-the-art methods. Moreover, we consider an example of the extracted summary from DUC 2002 using the proposed model (EMBAWEF) shown in the Appendix, **Fig. 5**.

**Table 6.** ROUGE scores for English on MultiLing2015 dataset

| Model | English | |
|---|---|---|
| | ROUGE -1 | ROUGE -2 |
| Worst | 37.17% | 9.93% |
| Best | 50.38% | 14.12% |
| C W2V (Rossiello, Basile, and Semeraro 2017) | 50.43% | 13.34% |
| Ours (EMBAWEF) | **51.59%** | **15.41%** |

# 6. Conclusion and future work

Traditional approaches to extract important information from a document rely heavily on human engineering features. In this paper, word embeddings are utilized to enhance the latent semantic analysis (LSA) input matrix weighting schemes. The proposed model, EMBAWEF, is used to compute the cell values for LSA input matrix. From which, three matrices will yield by applying singular value decomposition (SVD) on this matrix. The experimental results on DUC 2002 and DUC 2004 datasets show that the proposed models improve the performance of LSA algorithm in document summarization, especially EMBAWEF. The results also show the applicability of the model to extract the important sentences from the source effectively. The model achieves higher ROUGE scores than several well-known approaches. Although the new weighting schemes are evaluated on the document summarization task, it can be used in other information retrieval and NLP applications such as text similarity and web search. In the future work, we will explore other methods suitable for our problem; the deep-learning-based approach is a promising direction.

# ACKNOWLEDGEMENTS

We are grateful to the support of the National Natural Science Foundation of China (Grant No. 61379109, M1321007) and Science and Technology Plan of Hunan Province (Grant No. 2014GK2018 ,2016JC2011). We would like to thank the anonymous referees for their helpful comments and suggestions.

# Appendix

**Fig. 6** is a snapshot from the implementation of this work in python 3.6. The example shows one representative document from DUC2002 (D081A/AP891103-0200) along with its gold and system summaries. It demonstrates that the model EMBAEF performs a good performance identifying the key sentences in the document.

```
Python 3.6.0 |Anaconda 4.3.0 (64-bit)| (default, Dec 23 2016, 11:57:41) [MSC v.1900 64 bit (AMD64)]
Type "copyright", "credits" or "license" for more information.
2017-08-23 10:27:28,264 : INFO : loading projection weights from D:/Summarization/data/GoogleNews-vectors-negative300.bin
Reloaded modules: normalization, contractions, utilsfunctions, WV_Avg, LSAEmbeddingMatrixBuilder_V2
2017-08-23 10:29:33,627 : INFO : loaded (3000000, 300) matrix from D:/Summarization/data/GoogleNews-vectors-negative300.bin
Document Name: D081.P.AP891103-0200
Number of Sentences in The Document: 14

The Original Document:

{'D081.P.AP891103-0200': 'Thousands of miners in the northern Vorkuta region are expanding their
strike and some are blocking coal shipments the Soviet news media reported Friday. The miners are
demanding the government fulfill promises of improved living and job conditions. Soviet officials
have said the strikes could force fuel rationing during the Soviet Union s severe winter. Premier
Nikolai I. Ryzhkov introduced a bill in parliament that would increase pension benefits by about
percent and upgrade benefits for coal miners the official Tass news agency reported. The
Komsomolskaya Pravda youth newspaper in an article giving a sympathetic view of the workers
condition s in the Arctic Cirle coal mining districts said many miners do not see daylight for
months because of their underground work and the sun dips below the horizon in the winter. A
regional court in Vorkuta ruled that the latest round of strikes is illegal but did not impose any
penalties. The miners s unions said the decision will be appealed. Coal miners in the northern
region and the Ukraine struck for two weeks this summer but returned to work in July after
parliament passed a resolution promising reforms including improved social and economic conditions.
The miners say the government has renedged on its promises. The news media gave these accounts of
the latest strikes Workers at the Vorgashor mine in the Arctic Circle the largest mine in the
Vorkuta region continued their strike for an eighth day according to Tass. Komsomolskaya Pravda
said night shift workers walked out at three mines in Vorkuta and at another one the miners were
still working but were preventing the coal from being shipped outside the region. In the Ukraine s
Donetsk Coal Basin the nation s major coal producing area miner representatives were discussing
another strike. Tens of thousands of miners walked off the job for two hours Wednesday. Tass said
that in Donetsk the miners gathered in front of the House of Unions to demand that the government
set a deadline for implementing reforms. '}

The Gold Summary

Miners in the northern Vorkuta region are expanding their strike and blocking coal shipments,
demanding the government fulfill promises of improved living and job conditions.  In the Arctic
Circle coal-mining districts, miners do not see daylight for months because of their underground
work and the sun dips below the horizon in winter.
Coal miners in northern Ukraine struck for two weeks this summer but returned to work after a
resolution was passed promising reforms.  The miners say the government has reneged on its
promises.
A bill introduced by Premier Ryzhkov would increase pension benefits by 40% and upgrade
miners' benefits./n

The Generated Summary Using EMBAWEFLSA Model:

The Komsomolskaya Pravda youth newspaper in an article giving a sympathetic view of the workers
condition s in the Arctic Cirle coal mining districts said many miners do not see daylight for
months because of their underground work and the sun dips below the horizon in the winter.
Coal miners in the northern region and the Ukraine struck for two weeks this summer but returned to
work in July after parliament passed a resolution promising reforms including improved social and
economic conditions.
The news media gave these accounts of the latest strikes Workers at the Vorgashor mine in the
Arctic Circle the largest mine in the Vorkuta region continued their strike for an eighth day
according to Tass.
```

**Fig. 5.** Example document, gold summary and system summary from DUC 2002 using EMBAWEF Model and word2vec